\documentclass{article}

\usepackage{microtype}
\usepackage{graphicx}
\usepackage{subfigure}
\usepackage{booktabs} %
\usepackage{enumitem} %

\usepackage{hyperref}

\usepackage[accepted]{icml}

\usepackage{amsmath}
\usepackage{amssymb}
\usepackage{mathtools}
\usepackage{amsthm}

\usepackage[capitalize,noabbrev]{cleveref}

\theoremstyle{plain}

\theoremstyle{definition}

\theoremstyle{remark}

\usepackage[textsize=tiny]{todonotes}

\icmltitlerunning{Modeling Code: Is Text All You Need?}

\begin{document}

\twocolumn[
\icmltitle{Modeling Code: Is Text All You Need?}

\icmlsetsymbol{equal}{*}

\begin{icmlauthorlist}
\icmlauthor{Daniel Nichols}{umd}
\icmlauthor{Konstantinos Parasyris}{llnl}
\icmlauthor{Harshitha Menon}{llnl}
\icmlauthor{Brian R.~Bartoldson}{llnl}
\icmlauthor{Giorgis Georgakoudis}{llnl}
\icmlauthor{Tal Ben-Nun}{llnl}
\icmlauthor{Abhinav Bhatele}{umd}
\end{icmlauthorlist}

\icmlaffiliation{umd}{Department of Computer Science, University of Maryland, College Park, USA}
\icmlaffiliation{llnl}{Center for Applied Scientific Computing, Lawrence Livermore National Laboratory, Livermore, USA}

\icmlcorrespondingauthor{Daniel Nichols}{dnicho@umd.edu}
\icmlcorrespondingauthor{Abhinav Bhatele}{bhatele@cs.umd.edu}

\icmlkeywords{Machine Learning, ICML}

\vskip 0.3in
]

\printAffiliationsAndNotice{}  %

\begin{abstract}
Code LLMs have become extremely popular recently for modeling source code
across a variety of tasks, such as generation, translation, and summarization.
However, transformer-based models are limited in their capabilities to reason
through structured, analytical properties of code, such as control and data
flow. Previous work has explored the modeling of these properties with
structured data and graph neural networks. However, these approaches lack the
generative capabilities and scale of modern LLMs. In this work, we introduce a
novel approach to combine the strengths of modeling both code as text and more
structured forms.

\end{abstract}

\section{Introduction}
\label{sec:intro}
Modern large language models (LLMs) have shown remarkable promise in
understanding and generating source code across tasks such as completion,
translation, and summarization. By leveraging massive amounts of code data
available, these models have demonstrated strong capabilities in modeling
lexical and syntactic aspects of code. However, they tend to struggle in
reasoning about more complex, structured properties of code.

One prominent shortcoming of current code LLMs is their difficulty in accurately
modeling structured, analytical properties of code, such as control and data
flow. These types of relationships are best captured by graph-based models and
the sequential representations of transformers are not well-suited for this
task. Without explicit awareness of this structure, LLMs tend to rely heavily on
surface patterns in text and are prone to errors in tasks requiring
understanding of code structural properties. An example consequence of this
limitation is the lack of symmetry under semantic-preserving code
transformations. Two versions of a code that do the exact same thing may lead to
distant internal representations in an LLM, while the static graph-based
representation of the code would be identical. This would require more
complicated modeling on top of the LLM's internal representation than with a
graph-based approach.

Fixing this limitation of current code LLMs is valuable for many downstream
tasks that rely on code analysis, including testing, debugging, security
vulnerability detection, and performance optimization. If LLMs can robustly
capture and utilize the structured properties of code, they can better assist
developers with high-level reasoning, automatically suggesting safer or more
efficient code. This would in turn reduce the development time and effort to
produce high-quality software.

While it is important to develop models that merge the benefits of code LLMs
and graph reasoning, it is also non-trivial. Graph neural networks are
effective at modeling structured code information, but they lack the
large-scale generative abilities of LLMs. Conversely, LLMs do not readily
accommodate graph representations, and naive attempts to encode graphs into
text often do not scale well or lose important structural information. Ideally,
a new approach should build off of the modeling capacity and generative
capabilities of modern LLMS and ameliorate them with better structured
reasoning around code.

In this paper, we propose a solution that integrates graph-based reasoning into
LLMs using a GNN soft prompting approach. Our method learns how to encode
structured code representations into prompts that can be consumed by powerful
pre-trained LLMs. To accomplish this, we propose a novel graph representation
of LLVM intermediate representation (IR) that can be learned to be mapped into
a language model's embedding space.  By bridging key information from
graph-based analyses directly into the model's latent space, we preserve the
fidelity of structured reasoning, and achieve the flexibility and scale of
modern LLMs.

\section{Background}
\label{sec:background}
Below, we present the background on various structured graph
representations of code and soft prompting for LLMs.

\subsection{Structured Code Representations}
\label{sec:background:structured-code-representations}

When employing ML techniques to model code properties it is often beneficial to
represent the code in a structured form such as abstract syntax tree (AST) or
control flow graph (CFG). The nature of code generally permits the construction
of such structured representations and they are often more informative than the
raw text and allow models to predict code properties with higher accuracy and
less parameters.

The earliest approaches to modeling code with structured properties combined
ASTs with statistical features of the code to predict properties such as the
underlying algorithm~\cite{mou:aaai2016, allamanis2018learning}. These
approaches used graph neural networks (GNNs) and custom tree convolutional
networks to learn from the ASTs. However, more recent state-of-the-art
approaches have had better success modeling graph constructions of intermediate
representations (IR), such as LLVM IR~\cite{pmlr-v139-cummins21a,
jamsaz2024perfograph}. The former, ProGraML~\cite{pmlr-v139-cummins21a},
constructs a graph where each node is an IR instruction and edges represent
control, call, and data flow in the program. This representation, paired with a
GNN, is effective at tasks such as algorithm classification and relative
performance prediction. The latter, Perfograph~\cite{jamsaz2024perfograph}, is
an extension of this work that improves on the node embeddings in the graph
construction. Other works have explored similar constructions of CFGs for
security vulnerability modeling~\cite{steenhoek2024dataflow}.

\subsection{Soft Prompting}
\label{sec:background:soft-prompting}

Often when utilizing LLMs prompt engineering is necessary to find the right
inputs to the model to achieve the desired output. However, this can be
difficult as it is not systematic and must be done by hand. One could instead
fine-tune a model for the desired tasks to enforce a stricter objective,
however, this itself is time-consuming and requires a large amount of compute
resources. Soft prompting is a technique that lies in between the two approaches
where we instead learn how to optimally map inputs into the input space of an
existing LLM. This allows us to use an existing LLM without the need to
fine-tune it or hand engineer prompts. Soft prompting has been shown to be
effective at learning how to better provide information to LLMs and condition
them for particular downstream
tasks~\cite{lester2021powerscaleparameterefficientprompt,
li2021prefixtuningoptimizingcontinuousprompts, liu2023gptunderstands,
wang2023multitaskprompttuningenables,
blau2024contextawareprompttuningadvancing}. Building on top of these works,
recent papers have shown that GNN-based soft prompting can be effective at
reasoning through structured properties of graphs~\cite{Liu_2024,
perozzi2024letgraphtalkingencoding}.

\section{Collecting IR Data at Scale}
\label{sec:data-collection}
In order to learn from structured representations of code, we must first
collect such data at a large enough scale for training purposes. In this
section, we describe the process of collecting code and IR data at scale.

\subsection{Collecting Pairs of Source Code and LLVM IR}
\label{sec:data:ir}

While large code datasets already exist~\cite{Kocetkov2022TheStack,
lozhkov2024starcoder}, they are limited to source code. When any meta-data is
included, it is text data such as comments, documentation, git commits,
schemas, etc. These datasets do not contain structured information such as
abstract syntax trees (ASTs) or intermediate representation (IR) from compilers
such as LLVM. While some of this data can be extracted from the source code, IR
requires compilation, and hence, is non-trivial to collect from arbitrary
codebases at scale.  One large project, the ComPile
dataset~\cite{grossman2024compile}, contains a large amount of LLVM IR
bytecode, but does not contain the source code paired with it. To jointly learn
from the source and IR, both of these are required as paired data.

To collect this data, we build off of the work in Grossman et
al.~\cite{grossman2024compile} and use the ComPile dataset as a starting point.
We re-compile the C/C++ code in this dataset to LLVM IR and extract the
corresponding source code from the metadata in the Spack package
manager~\cite{gamblin:sc15}. Furthermore, we collect the IR data using LLVM 16,
with a custom version of the llvmlite library, to retrieve better data and
enable more transformations downstream.  The IR is collected without
optimizations as these can be applied with transformations using the {\tt opt}
tool in LLVM later. The final dataset totals approximately 2 million files of
C/C++.

\subsection{Collecting Synthetic Data}\label{sec:data:synthetic}

The dataset described above contains real-world, complete code files
alongside their compiled LLVM IR. To expand this dataset, we further collect
question and answer pairs from the code and IR. Each sample in this dataset
is a quadruple of the form $(\text{source code}, \text{IR}, \text{question},
\text{answer})$. Questions and answers are generated synthetically using an
LLM, namely GPT-4o~\cite{openai2024gpt4ocard_short_author}, in a similar format
to the data collection in~\cite{wei2023magicoder}. We provide a code
snippet to the LLM and a random text snippet from the The Stack dataset
~\cite{lozhkov2024starcoder} to use as inspiration. Given the code and
inspiration, the LLM is tasked with generating a question and answer pair about
the snippet of code. This structure is particularly useful for the task of
mapping IR and question pairs to answers, i.e. $\text{IR} \times \text{Q}
\mapsto \text{A}$.

We collect another CodeQA dataset where question answer pairs are synthetically
generated using an LLM and the answers are somewhere in the context of the
source code. For example, a sample has source code, a question, and an answer
where the answer is simply a location (or locations) in the source code that
answer the question. This dataset is useful for testing how well a model can
reason through structural properties of the code. If the model can generate
correct responses to questions about structural properties of the code, it is
likely that it has learned to model the code at a deeper level than merely the
source code as text.

\section{An Improved Structured Graph Format}
\label{sec:improving-structured-graph}
Next, we present an enhanced structured graph format designed to
represent LLVM Intermediate Representation (LLVM IR) constructs with greater
fidelity and granularity. Building upon prior work such as ProGraML, our
approach incorporates additional node and edge types, enabling a richer modeling
of the elements present in LLVM IR. The goal of these enhancements is to
improve the expressivity of the graph representation and better capture the
underlying semantics and dataflow relationships in the IR.

\subsection{Design of the IRGraph Format}\label{sec:graph-format:design}

The graph format is inspired by similar representations of code, such as
ProGraML and PerfoGraph, however, it uses a finer granularity to split IR
statements into the graph. It further adds more node and edge types to better
distinguish information on the graph. The final graph format, IRGraph, has six
node types and eight edge types, which are described below.

\begin{itemize}[align=left, noitemsep]
    \item[\textbf{Node Types}]
    \item[\textit{Value:}] Represents individual LLVM IR values, such as variables or constants.
    \item[\textit{Type:}] Encodes type information, such as integer or floating-point types, associated with LLVM IR values.
    \item[\textit{Size:}] Models container sizes or dimensionality information derived from type properties.
    \item[\textit{Module:}] Represents the LLVM IR module as a global context for values and functions.
    \item[\textit{Attributes:}] Captures function and argument attributes, including linkage, visibility, and calling conventions.
    \item[\textit{Instruction:}] Represents individual instructions in the LLVM IR, including their operation codes and alignment information.
\end{itemize}

\begin{itemize}[align=left, noitemsep]
    \item[\textbf{Edge Types}]
    \item[\textit{Type:}] Connects a value to its associated type (value $\rightarrow$ type).
    \item[\textit{Dataflow:}] Captures data dependencies, including definitions (instruction $\rightarrow$ value) and uses (value $\rightarrow$ instruction).
    \item[\textit{Attribute:}] Links values to their attributes (value $\rightarrow$ attribute).
    \item[\textit{CFG:}] Represents control flow between instructions (instruction $\rightarrow$ instruction).
    \item[\textit{Size:}] Maps types to their associated sizes (type $\rightarrow$ size).
    \item[\textit{Symbol:}] Connects the module to global values (module $\rightarrow$ value).
    \item[\textit{Includes:}] Connects contained types (type $\rightarrow$ type).
    \item[\textit{Contains:}] Connects constants and global variables to operands and initializers (value $\rightarrow$ value).
\end{itemize}

This structure provides more comprehensive detail about the LLVM IR code beyond
control and data flow, which are typically the only focus of graph code
representations such as ProGraML and PerfoGraph. By incorporating more granular
details, the representation is better able to capture nuanced relationships
between different elements of the LLVM IR code. For example, attributes,
symbols, and size information are crucial to understanding performance
properties of code, which is a common task for graph-based code representations,
yet none of the related works incorporated these details or only encapsulated
them in a limited manner.

\subsection{Graph Construction Process}
\label{sec:graph-format:construction}

The construction of the IRGraph representation begins with parsing LLVM
Intermediate Representation (IR) files using LLVM 16, ensuring compatibility
with the latest features and constructs of the language. A Python script and
llvmlite Python bindings are used to extract nodes and edges, adhering to the
structure described in the previous section. We updated portions of the llvmlite
and numba libraries to support the parsing the needed LLVM IR constructs. Graphs
are constructed and stored as PyTorch Geometric HeteroData objects.

We first construct nodes for values, types, instructions, attributes, and the
module itself in the LLVM IR. Each node is assigned a unique feature vector that
encodes its type-specific properties. For instance, value nodes include
information about the kind of value they represent (e.g., constants, variables),
while instruction nodes capture details such as the opcode. Type nodes encode
structural details, including whether the type is scalar, vector, or array, and
any associated dimensions.

Once the nodes are initialized, relationships between these components are
identified and represented as edges. We first connect edges to the value nodes.
These are the type, dataflow, and attribute edges. Type nodes are further
connected to size nodes encoding the size of the type. Values are finally
connected to the module they reside in using symbol edges. These are constructed
as undirected edges to allow information to flow back to module nodes during GNN
training. To encapsulate execution flow we connect instruction nodes in the
graph using control flow edges to represent the execution order of instructions.
Finally, type nodes are connected to other type nodes using includes edges to
represent type hierarchies, and value nodes are connected to other value nodes
using contains edges to represent the relationship between constants and global
variables and their operands and initializers.

One major distinction between our approach and prior work is the ability to
represent entire compilation units. Previous works, such as ProGraML, focus on
single functions and are not equipped to handle the entire IR module. Our
approach is able to contextualize all of the information available in an IR file
and represent it in the graph. This is a significant advantage as the code we
desire to model is rarely a single function, but rather an entire program or
compilation unit. For each entire program, we can construct an IR graph and
store it as a PyTorch Geometric HeteroData object that encodes six node types
and eight edge types. Furthermore, it stores feature vectors for each node in
the graph based on the node type.

\section{Learning From Text and Graphs}
\label{sec:methodology}
We now present our methodology for integrating graph-based modeling
with LLMs through a GNN soft prompting approach. Our method learns how to encode
structured code representations into prompts that can be input into existing
pre-trained LLMs. By bridging key information from graph-based analyses directly
into the model's latent space, we achieve the flexibility and scale of LLMs
while preserving the fidelity of structured modeling. We propose two new
representations, the new graph format plus GNN, IRGraph, and the graph plus LLM
representation, IRCoder.

\subsection{IRGraph: Graph Masked Pre-Training}\label{sec:methodology:graph-masked-pretraining}

The first component of our architecture is a graph neural network (GNN) used to
model the novel graph representation proposed in this work. The GNN is a
heterogeneous GNN as the graph has multiple node and edge types. The nodes
represent literals and statements in the LLVM IR, while the edges represent
various relationships between them, such as caller-callee, data flow, etc. To
learn from these heterogeneous graphs, we use a two layer GCN with message
passing along each unique edge type.

Each layer in the model is a collection of message passing functions $\{f_1,
f_2, \ldots, f_n\}$, where each function $f_i$ is a function that passes
messages for edge type $i$. We utilize standard graph
convolutions~\cite{KipfW16} for message passing along individual
edge types. For messages that share a destination node type, i.e. IR literals,
we accumulate the passed messages. Since the {\tt module} node type only has
outgoing edges, we reciprocate the incoming edges to allow for information to be
propagated to the {\tt module} nodes.

This model outputs tensors of shape $B \times \lvert V \rvert \times h_i$, where
$B$ is the batch size, $\lvert V \rvert$ is the number of nodes in the graph,
and $h_i$ is the node output embedding dimension. For graph level embeddings,
we further aggregate along the node dimension using mean pooling. The resulting
$B \times h_i$ tensor is then projected to the same dimension as the LLM's
token embedding space, $E$, using a linear layer.

This model is initially pre-trained using masked pre-training on the node
values. We mask out a random subset of the node values and predict them using
the other node values in the graph. Each training iteration a random node type
is selected to be masked out and predicted. This pre-training step is crucial
for learning the structure of the graph and the relationships between the nodes.
Furthermore, it allows us to make use of the large amount of unlabeled data we
have available in the IR dataset.

\subsection{IRCoder: Soft Prompt Fine-Tuning}\label{sec:methodology:soft-prompt-finetuning}

To align the graph-based encoder with a pretrained language model, we introduce
a small set of learnable parameters acting as a prompt. Specifically, we prepend
trainable vectors to the text embeddings of the LLM. To accomplish this, we
compute one graph embedding $\mathcal{G}$ of dimension $E$ and $\lvert V \rvert$
node embeddings $\mathcal{V}_i$ of dimension $E$ of a particular source code. We
then tokenize and embed the source code to $\mathcal{T}_i$ using the LLM's
tokenizer. Using the graph embeddings and token embeddings, we construct the
following input to the LLM:
$$ \left[
    BOS, 
    \mathcal{G}, 
    \mathcal{V}_1, \ldots, \mathcal{V}_{\lvert V \rvert},
    \mathcal{T}_1,
    \ldots,
    EOS
\right] $$

The GNN outputs are combined with the token embeddings, allowing the model to
attend to structural features while generating or classifying text. During
fine-tuning we freeze the LLM weights and only update the GNN weights using
backpropagation from the loss. This drastically reduces the compute, both memory
and time, required to fine-tune the model. It further enables us to make use of
existing LLMs without the need for full pre-training, which can be expensive and
time consuming. One limitation of this approach is that it increases the use of
the model's context and may lead to requiring larger contexts to reasonably fit
the graph in the prompt. Reducing the number of graph tokens is a potential
future research direction, however, we rely on the recent scaling trends in
model context windows to primarily mitigate this issue.

\section{Experiments}
\label{sec:experiments}
In this section we highlight the benchmarks used for evaluating our approach,
alongside the models and training procedures employed in our experiments.

\subsection{Benchmarks}\label{sec:experiments:benchmarks}

\noindent\textit{DevMap Dataset:} 
The DevMap, previously used in~\cite{pmlr-v139-cummins21a}, dataset consists of
OpenCL kernels annotated with their performance metrics on different hardware
platforms, including a CPU, an NVIDIA GPU, and an AMD GPU. The primary task is
to predict whether a given kernel will perform better on the CPU or the GPU
based on its source code and intermediate representation (IR) graph. This task
is divided into two subtasks: one focusing on predicting performance on the
NVIDIA GPU and the other on the AMD GPU. The dataset provides a comprehensive
benchmark for evaluating the effectiveness of our approach in understanding and
predicting hardware-specific performance characteristics of OpenCL kernels. This
benchmark reveals both source code (the OpenCL) and an LLVM IR graph making it
an ideal test case for our approach.

\noindent\textit{Algorithm Classification:}
The POJ-104 dataset~\cite{mou2016aaai} is used for the algorithm classification
task. This dataset contains 104 different algorithm classes, each represented by
multiple C++ programs. Each sample in the dataset includes the complete source
code of an algorithm and its testing framework. From this we can compile and
generate LLVM IR. The objective is to predict the algorithm class based on the
provided source code and IR. This benchmark allows us to evaluate the capability
of our approach in accurately classifying algorithms, demonstrating its
effectiveness in understanding and distinguishing between different types of
algorithms based on their code and IR features.

\noindent\textit{Vulnerability Detection:}
The Juliet test suite~\cite{juliet} is utilized for the vulnerability detection
task. This dataset comprises approximately 100,000 C++ samples, each having a
version with and without a security vulnerability. The task is to predict
whether a given code sample contains a vulnerability or not. We use pair-wise
accuracy as the evaluation metric, which means the model must correctly identify
both the vulnerable and non-vulnerable versions of each sample. Previous
literature has demonstrated that pair-wise accuracy is a more realistic metric
for evaluating vulnerability detection models. For each sample, the model is
provided with the source code and its corresponding IR graph. This benchmark is
critical for assessing the model's ability to detect security vulnerabilities in
code.

\noindent\textit{Code Translation:}
The ParEval benchmark~\cite{nichols:hpdc2024} is employed for the code
translation task, which involves translating code from one parallel programming
model to another. Specifically, we evaluate translations from sequential code to
OpenMP, sequential code to MPI, and OpenMP code to CUDA. Each sample in this
benchmark consists of a single C++ kernel, and the model's objective is to
predict the translated code. The translations are scored based on functional
correctness. The ParEval benchmark includes 60 problems for each execution
model, covering 12 distinct problem types. This benchmark provides a rigorous
test for assessing the ability of our approach to accurately translate code
between different parallel programming models, ensuring functional correctness
and performance. We use this benchmark as it enables us to evaluate the
generative performance of the combined model; generative tasks are a key
strength of the model as purely graph-based approaches are not well-suited to
generative tasks.

\subsection{Models and Training}\label{sec:experiments:models}

We begin by evaluating the effectiveness of our new graph structure. To identify
the optimal architecture for each task, we conduct an extensive architecture
search across various GNN architectures, including Graph GCNs, Graph Attention
Networks (GATs), and GraphSAGE. We enumerate different configurations of these
architectures, varying the number of layers and hidden dimensions. This
architecture search is done for each benchmark to determine the best overall
configuration. Completing this search first enables us to use the best possible
embeddings in the subsequent soft-prompting setup.

After identifying the best architecture, we proceed with a two-stage training
process: masked pretraining followed by prompt fine-tuning. During the masked
pretraining phase, we use the best performing GNN architecture and do masked
node prediction to pre-train it on the large unlabeled IR dataset. This gives
us a pre-trained graph model that can be further fine-tuned for soft-prompting.

Once pre-training is complete, we fine-tune the model using prompt-based
learning. In this step, we use the pre-trained GNN and LLM to fine-tune the GNN
further for soft-prompting. This fine-tuning is done over the large IR plus
source code dataset, in addition to the synthetic datasets. We use cross entropy
loss for next token prediction and only update the weights for the GNN with the
backpropagated gradients. Fine-tuning is completed for one full epoch of the
combined training datasets with a learning rate of $1e-4$ and the
AdamW~\cite{adamw} optimizer.

\section{Results}
\label{sec:results}
In this section we highlight the results of each of the benchmarks and 
further ablate portions of the IR graph representation.

\subsection{Device Mapping}\label{sec:results:device_mapping}

\begin{figure}[ht]
    \centering\includegraphics[width=\linewidth]{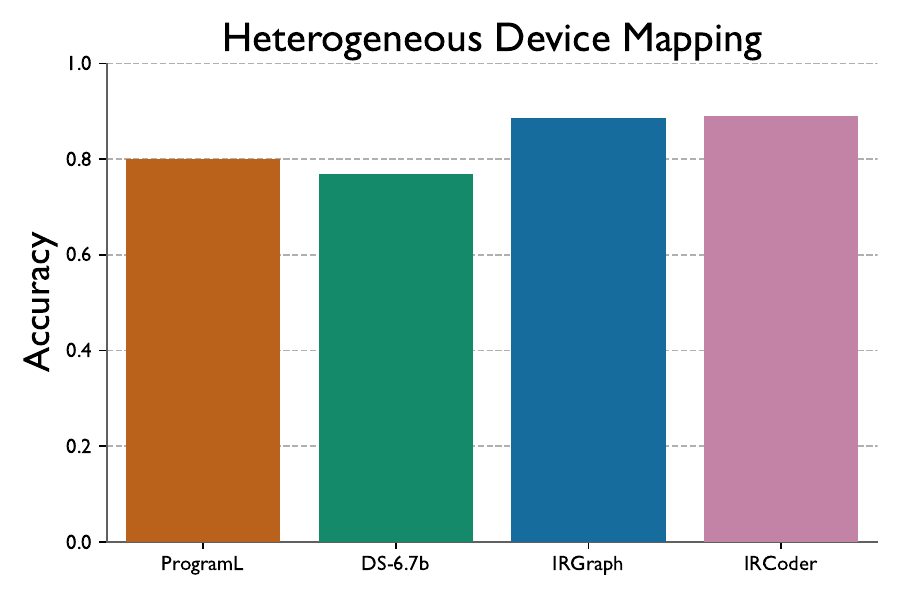}
    \caption{Accuracy scores from the DevMap benchmark. Both of the proposed
    representations outperform the respective baselines. The IRGraph graph
    representation improves on the ProGraML graph model, while the IRCoder
    language model builds on the graph to improve the language model.
    \label{fig:devmap}
    }
\end{figure}

\Cref{fig:devmap} shows the results of the different representations on the
DevMap benchmark. We compare the proposed IRGraph and IRCoder representations
against baseline IR graph and LLM models, namely ProGraML and
Deepseek-Coder-6.7b. The results are shown for each of the models fine-tuned on
the DevMap train set and evaluated on the DevMap test set. The LLMs are
fine-tuned with a classification head instead of a language modeling head. The
proposed representations both outperform the baselines. We see that just the
IRGraph representation improves on the ProGraML graph model in predicting the
faster device, CPU or GPU. Furthermore, the IRCoder model improves on the
baseline language model demonstrating the capability of the soft-prompting
technique to improve the modeling capacity of LLMs.

\subsection{Algorithm Classification}\label{sec:results:algorithm_classification}

\begin{figure}[ht]
    \centering\includegraphics[width=\linewidth]{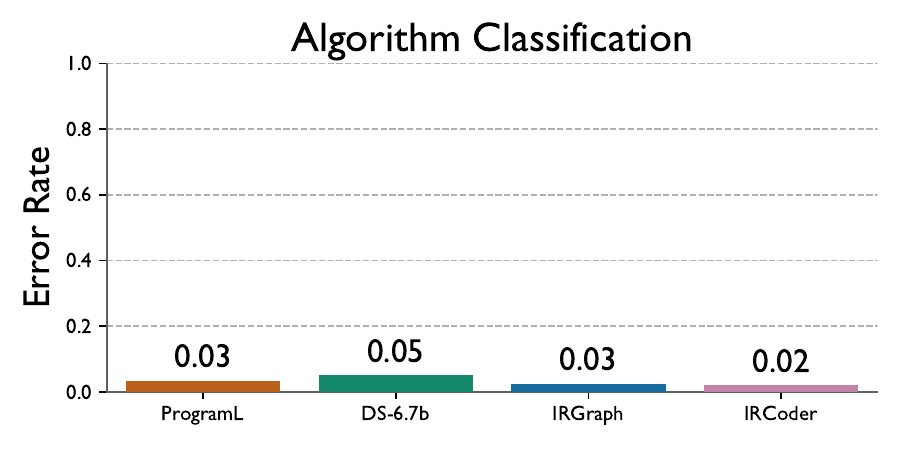}
    \caption{Error rate scores from the POJ-104 benchmark. All representations
    are strong at this task. The IRGraph representation scores the same as
    ProGraML while the IRCoder representation outperforms the Deepseek-Coder
    baseline.
    \label{fig:poj104}
    }
\end{figure}

The results on the POJ-104 benchmark are shown in \Cref{fig:poj104}. Here we
show the error rate for each representation on the task of classifying the code
samples into one of 104 algorithm classes. The proposed IRGraph representation
performs on par with the ProGraML graph model, however, the Deepseek-Coder
language model performs worse. The IRCoder model is able to combine the benefits
of the LLM and IRGraph representations to outperform all baselines. These 
trends are consistent with the DevMap benchmark results.

\subsection{Vulnerability Detection}\label{sec:results:vulnerability_detection}

\begin{figure}[ht]
    \centering\includegraphics[width=\linewidth]{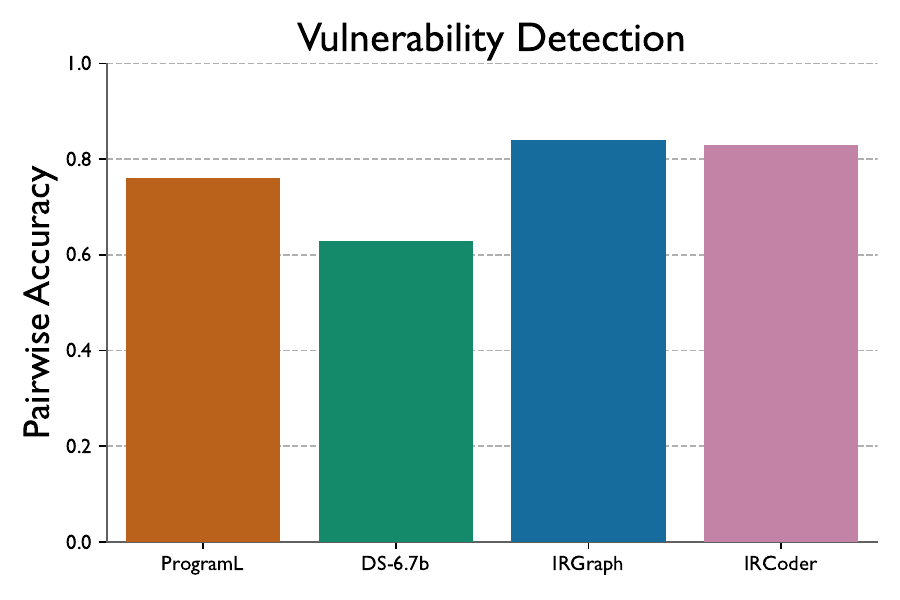}
    \caption{Pair-wise accuracy scores from the Juliet benchmark. Pairwise
    accuracy, where a correct prediction requires both the vulnerable and
    non-vulnerable versions of a sample to be correctly classified, is used as
    the evaluation metric. The IRGraph and IRCoder representations outperform
    the baselines.
    \label{fig:vuln-detect}
    }
\end{figure}

The final classification benchmark, vulnerability detection, is shown in
\Cref{fig:vuln-detect}. In this benchmark the model is shown a version of a
C/C++ code from the Juliet dataset and is tasked with predicting if it is
vulnerable or not. Following trends in recent literature we measure pair-wise
accuracy where a correct prediction requires both the vulnerable and
non-vulnerable versions of a sample to be correctly classified. 

We observe similar trends for vulnerability detection to the other two
classification benchmarks. The IRGraph and IRCoder models outperform the
baseline models. However, unlike the other two benchmarks we see slightly worse
pair-wise accuracy for IRCoder than IRGraph, albeit the difference is small.
Altogether, these results demonstrate the effectiveness of the proposed
representations for modeling code properties. Furthermore, they demonstrate the
effectiveness of incorporating structural information into language models for
code understanding tasks.

\subsection{Code Translation}\label{sec:results:code_translation}

\begin{figure}[ht]
    \centering\includegraphics[width=\linewidth]{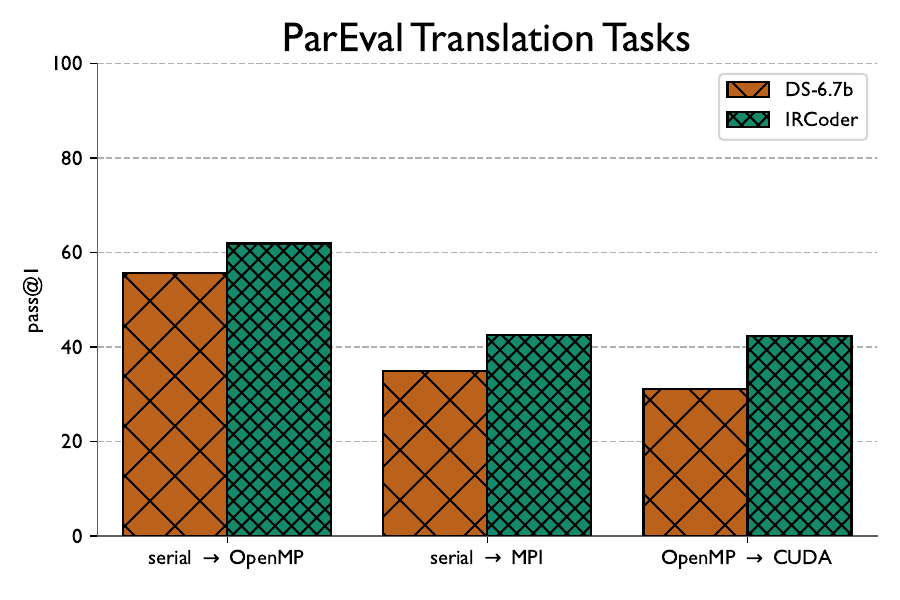}
    \caption{pass@1 scores from the ParEval benchmark comparing Deepseek-Coder
    and IRCoder. The IRCoder model is better able to translate code when
    provided with the IR graph during translation. The most pronounced
    improvement is for the OpenMP to CUDA translation.
    \label{fig:pareval}
    }
\end{figure}

The results on the ParEval benchmark are shown in \Cref{fig:pareval}. Here the
base LLM, Deepseek-Coder, is given a correct implementation of a function in one
parallel programming model and asked to generate the equivalent implementation
in another model. The IRCoder model uses the same setup except with the IR graph
of the source implementation provided in addition. This gives the model more
context about the structure and properties of the code it is attempting to
translate. Pass@1 scores are reported demonstrating the functional correctness
of the generated code.

We observe a substantial increase in correctness of the translated code between
the base LLM and IRCoder. This trend is observed across all translation tasks:
sequential to OpenMP, sequential to MPI, and OpenMP to CUDA. The most prominent
improvement is for OpenMP to CUDA translation. We hypothesize this is due to the
OpenMP and CUDA code being the most similar in terms of structure meaning the
LLM is able to better use the structural information available in the IR graph.

Improving generative tasks is a key strength of the proposed model. Even though
the GNN-only models are able to get comparable modeling performance with 
significantly less parameters, they are not suited to generative tasks. Being able
to reason about and understand deep code structures is an important capability 
for a code LLM to have as it allows for more accurate and contextually aware
code generation.

\subsection{IRGraph Ablation}\label{sec:results:irgraph_ablation}

\begin{figure}[ht]
    \centering\includegraphics[width=\linewidth]{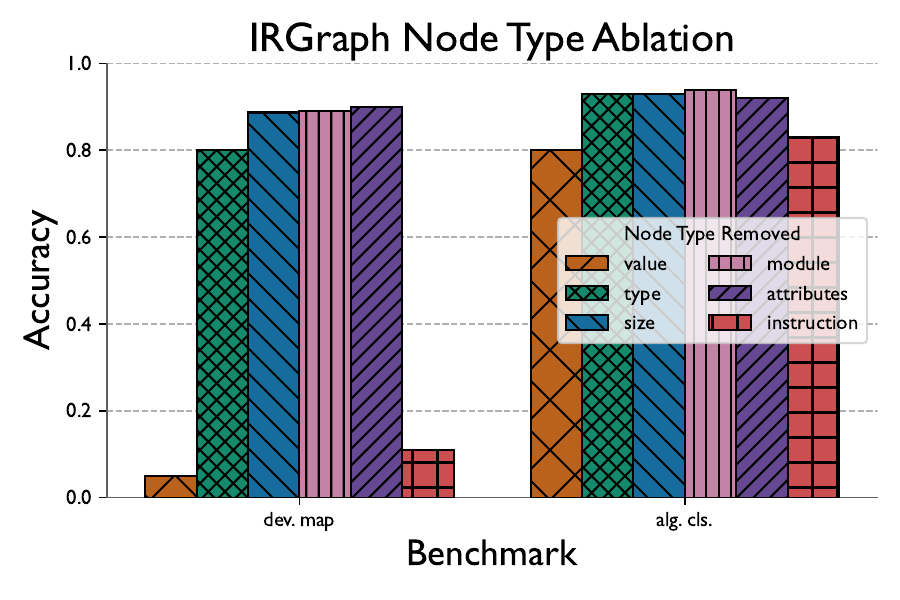}
    \caption{Ablation study by removing node types from the IRGraph
    representation. We see that value and instruction node types are the most
    important data points for modeling the IR. The IR attributes are the least
    important and only reduce the accuracy by less than 1\% when removed.
    \label{fig:irgraph-node-ablation}
    }
\end{figure}

To better understand the proposed graph representation, we perform ablation
studies to determine the importance of different node and edge types.
\Cref{fig:irgraph-node-ablation} shows the results of ablation studies on the
IRGraph representation. Results are collected by removing individual node and
edge types from the graphs and training the GNN model on the reduced graphs. In
some cases, this requires making edges bi-directional to prevent graphs where
node types have no incoming edges and information cannot be propagated to them
during message passing. Results are shown for both the DevMap and algorithm
classification benchmarks.

The results from the ablation study show that certain node types are more
critical for the model's performance. Specifically, the removal of value and
instruction nodes leads to a significant drop in accuracy. Conversely, the
removal of IR attribute nodes has a minimal impact on performance, suggesting
that these nodes contribute less to the overall model accuracy. The performance
drops are more significant for the device mapping benchmark, than for algorithm
classification, however, the trends are the same. This is likely due to the
device mapping task being a more complex task that requires more information
from the IR graph.

\begin{figure}[ht]
    \centering\includegraphics[width=\linewidth]{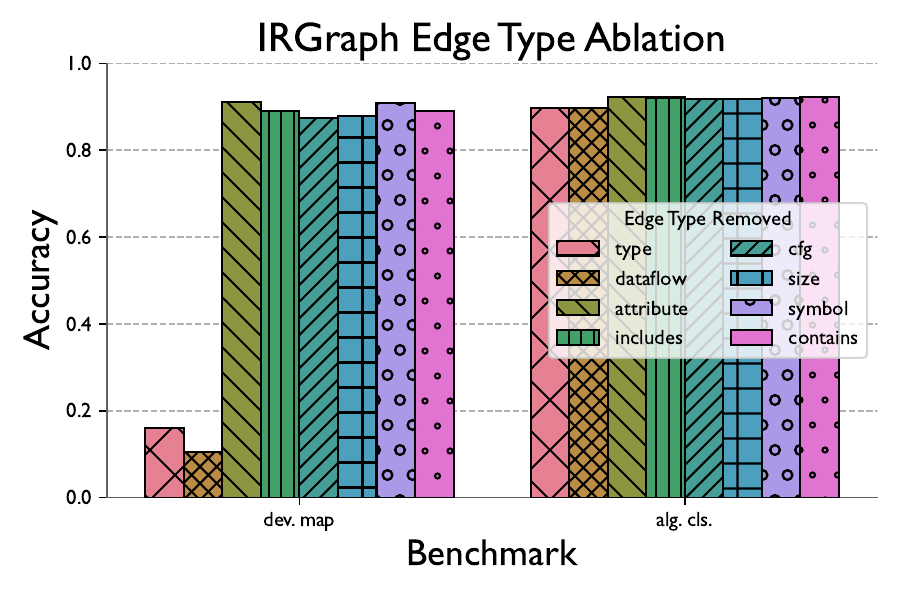}
    \caption{Ablation study by removing edge types from the IRGraph
    representation. We see that type and dataflow edges are the most important
    for the model's performance, while the other edge types have a minimal impact
    on accuracy.
    \label{fig:irgraph-edge-ablation}
    }
\end{figure}

A similar ablation study for the edge types is shown in
\Cref{fig:irgraph-edge-ablation}. Here we observe the type and dataflow edges
having the most significant impact on the model's performance. Surprisingly,
other edge types have a minimal impact on accuracy. Most notable is the removal
of control flow edges, which only reduces accuracy by 4 percentage points from
the full model for the DevMap benchmark. As with the node ablation study, the
trends are consistent, but less pronounced for the algorithm classification
benchmark. The type and dataflow edges lead to the largest drop in accuracy when
removed, but only by 3 percentage points for the algorithm classification
benchmark.

For both the node and edge ablation studies, we observe that the full IRGraph
representation outperforms all ablated versions, suggesting that all node and
edge types contribute to the model's performance. However, some types contribute
significantly less than others. As removing components never leads to better
performance there is no need to strip them from the final graph format.

\section{Related Work}
\label{sec:related-work}
Several other works have been proposed to address the problem of reasoning about
graphs in LLMs. Fatemi et al. introduced an approach for encoding graphs within
the prompt of an LLM~\cite{fatemi2023talklikegraphencoding}. This work unrolled
graphs directly into the prompt as text, allowing the model to directly take the
graph as input without any architectural modifications. A follow-up work took
this approach further by embedding graphs into the prompt using a
GNN~\cite{perozzi2024letgraphtalkingencoding}. Our work builds on top of the
latter approach by focusing on code, introducing a new graph structure that is
tailored to the IR of code, and expanding the type and amount of graph training
data. The latter study only trains the GNN using a small number of graph Q+A
samples, while we train on a large dataset of IR graphs. Furthermore, we
evaluate our approach on tasks beyond Q+A benchmarks.

Other works have explored integrating structured code information into a code
LLM. Two notable and related examples are
GraphCodeBERT~\cite{guo2021graphcodebertpretrainingcoderepresentations} and
UniXcoder~\cite{guo2022unixcoderunifiedcrossmodalpretraining}. Both works are
based on the BERT architecture and use ASTs and/or dataflow graphs alongside a
novel attention masking scheme to weight tokens based on the graph structure.
This approach is, however, limited in that it cannot encode complex graph
structures in the token sequence and only changes how existing code tokens are
interpreted. It is further limited by the need to unroll the AST into the token
sequence. This consumes valuable context space and increases the compute
requirements of the model's attention mechanism.

\section{Conclusion}
\label{sec:conclusion}
In this work, we introduced a novel approach for integrating structured
graph-based representations into LLMs through GNN-based soft prompting. By
leveraging LLVM IR and constructing enhanced program graphs, we demonstrated
that structured code representations can be effectively embedded into LLMs,
improving their ability to reason about control flow, data flow, and other
analytical properties of source code.

We accomplished this by first introducing a new graph structure of LLVM IR to
represent programs. We then demonstrated how to embed these graphs into an LLM's
embedding space using a GNN and soft prompting. Using this approach, we trained
a combined IR graph and text model, and evaluated it on several code modeling
and generation benchmarks. It was shown that the model outperforms the baseline
LLM and graph models on all benchmarks, demonstrating the effectiveness of our
approach. This work opens up new directions for combining structured and
unstructured representations in code intelligence tasks, and we believe that
this approach will be beneficial for a wide range of code understanding tasks in
the future.

\section*{Acknowledgements}
This work was performed under the auspices of the U.S.~Department of Energy by
Lawrence Livermore National Laboratory (LLNL) under Contract DE-AC52-07NA27344
(LLNL-CONF-2003845). This work was supported in part by the LLNL LDRD Program
under Projects No.~23-ERD-022.

\section*{Impact Statement}

This paper presents work whose goal is to advance the field of 
Machine Learning. There are many potential societal consequences 
of our work, none which we feel must be specifically highlighted here.

\bibliography{bib/pssg,bib/cite}
\bibliographystyle{icml}

\end{document}